\documentclass[runningheads]{llncs}
\usepackage[T1]{fontenc}
\usepackage{graphicx}
\usepackage{color}
\usepackage{xspace}
\usepackage{multirow}
\usepackage{booktabs}
\usepackage{colortbl}
\usepackage{url}
\usepackage{tcolorbox}
\usepackage{wrapfig}
\usepackage{newtxtext} 

\usepackage{fancyvrb}
\usepackage{fvextra}
\usepackage[inline,shortlabels]{enumitem} 
\usepackage{todonotes}
\usepackage{calc}

\newcommand{\gpt}{GPT-3\xspace}

\newcommand{\PETD}{PET\xspace}
\newcommand{\fa}{f_{a}\xspace}
\newcommand{\fb}{f_{b}\xspace}
\newcommand{\num}{seven\xspace}
\newcommand{\raw}{\textsc{Raw}\xspace}
\newcommand{\defs}{\textsc{Defs}\xspace}
\newcommand{\few}{\textsc{2Shots}\xspace}
\newcommand{\both}{\textsc{Defs+2Shots}\xspace}

\begin{document}
\title{Leveraging pre-trained language models for conversational information seeking from text}
\titlerunning{Leveraging pre-trained language models for CIS from text}
%
\author{Patrizio Bellan, Mauro Dragoni, Chiara Ghidini}
\authorrunning{Bellan et al.}
%
\institute{Fondazione Bruno Kessler, Trento, Italy \\
\email{pbellan|dragoni|ghidini@fbk.eu}}
%
\maketitle              
\begin{abstract}


Recent advances in Natural Language Processing, and in particular on the construction of very large pre-trained language representation models, is opening up new perspectives on the construction of \textbf{conversational information seeking} (CIS) systems. In this paper we investigate the usage of in-context learning and pre-trained language representation models to address the problem of \emph{information extraction from process description documents}, in an incremental question and answering oriented fashion. 
In particular we investigate the usage of the native \gpt (Generative Pre-trained Transformer 3) model, together with two in-context learning customizations that inject conceptual definitions and a limited number of samples in a few shot-learning fashion. The results highlight the potential of the approach and the usefulness of the in-context learning customizations, which can substantially contribute to address the ``training data challenge'' of deep learning based NLP techniques the BPM field. It also highlight the challenge posed by control flow relations for which further training  needs to be devised.

\end{abstract}
%
%
%


\section{Introduction}
\label{sec:intro}

Textual descriptions of business processes, contained for example in Standard Operating Procedure (SOP) documents, are ubiquitous in organisations. While the goal of most of these textual descriptions is that of being an easy to understand and use set of documents, the actual exploitation of the information they contain is often hampered by the challenge of having to manually analyse unstructured information. 

\emph{Process (information) extraction from text} can be regarded as the specific problem of finding algorithmic functions that transform textual descriptions of processes into structured representation of different expressivity, up to the entire formal process model diagram. The ambiguous nature of natural language, the multiple possible writing styles, and the great variability of possible domains of application make this task extremely challenging. Indeed recent papers on this topic~\cite{Riefer16,Maqbool18,Aa18,DBLP:Bellan-arxiv} highlight that after more than ten years of research from the seminal work in \cite{Friedrich11}, process extraction from text is a task far from being resolved and further research in this direction is needed with the purpose of improving the quality of the process model generation. 
By looking at the state of the art in this field, most of the existing approaches rely on template and rule based approaches, which often lack the flexibility needed to fully cover the great variability of writing styles and process domains \cite{AckermannV13,derAaDiCiccio19,GoncalvesSB11,SawantRSPS14,Friedrich11,EpureMHDS15,HonkiszK018,Ferreira17,DBLP:conf/bpm/QuishpiCP20}. Few recent works \cite{Han2020,DBLP:conf/caise/0003W0LLZLW20} try to leverage modern approaches based on deep learning Natural Language Processing (NLP), but they (somehow ironically) restrict the format of the source text to a structured text~\cite{Han2020} or to sequential lists of tasks such as recipes or assembly instructions~\cite{DBLP:conf/caise/0003W0LLZLW20}, thus failing to address the challenge posed by real world business process descriptions. 

One of the problems of leveraging the potential of deep learning NLP is the \textbf{lack of the high quantities of carefully annotated data on textual descriptions} needed to make these techniques work, which newly available annotated datasets, such as \PETD \cite{DBLP:pet-dataset}, are not yet able to address. The problem is made even worst by the multi-perspective nature of process elements (activities, data objects, actors, resources, flow objects, and their mutual relations, among others), which require articulated set of annotation labels and the planning of laborious annotation campaigns. 

Recent advances in NLP, and in particular the availability of large pre-trained language representation models, and the introduction of novel in-context learning strategies that enable the customization of these models in a few shot (a.k.a. few examples) fashion \cite{Raffel2020,Brown2020}, is opening up new perspectives on the construction of \textbf{conversational information seeking} (CIS) systems. These systems are drawing a growing attention in both academia and industry and aim at supporting search and question answering (among other tasks) by means of multi-turn dialogs in written or spoken form. 


In this paper we explore the feasibility of using in-context learning over the pre-trained language models to perform process extraction from textual documents in an incremental question and answering manner. In particular we explore the feasibility of using the native \gpt  (Generative Pre-trained Transformer 3) model, and two customizations built by providing conceptual definitions of business process elements and a limited number of examples in a few shot learning fashion. The results highlight the potential of the in-context learning approach which can substantially address the ``training data challenge'' of deep learning based NLP techniques the BPM field. It also highlight the challenge posed by control flow relations for which further specific training methodologies need to be devised. We also exploit the conversational nature of \gpt to investigate how to perform process extraction via incremental question answering. 
This can pave the way to the design of flexible pipelines which can seek information in an incremental manner, similarly to what happens with human experts.

The paper is structured as follows. We first provide some background on process extraction from text and in-context learning (Section~\ref{sec:background}); then we describe our approach and its empirical investigation using annotated texts from the \PETD \cite{DBLP:pet-dataset} dataset (Sections \ref{sec:approach} and \ref{sec:evaluation}); finally we report insights and lessons learned (Section \ref{sec:discussion}), related works (Section \ref{sec:related}) and concluding remarks.    

\section{Background}
\label{sec:background}

In this section we provide the main concepts needed for understanding the remainder of the paper.

\subsection{Process extraction from text}
\label{ssec:pet}
\emph{Process information extraction from text} can be regarded as the specific problem of finding algorithmic functions that transform textual descriptions of processes into structured representation of different expressivity, up to the entire formal process model diagram. 


\begin{figure}[tb]  
    \centering
    \includegraphics[width = \textwidth]{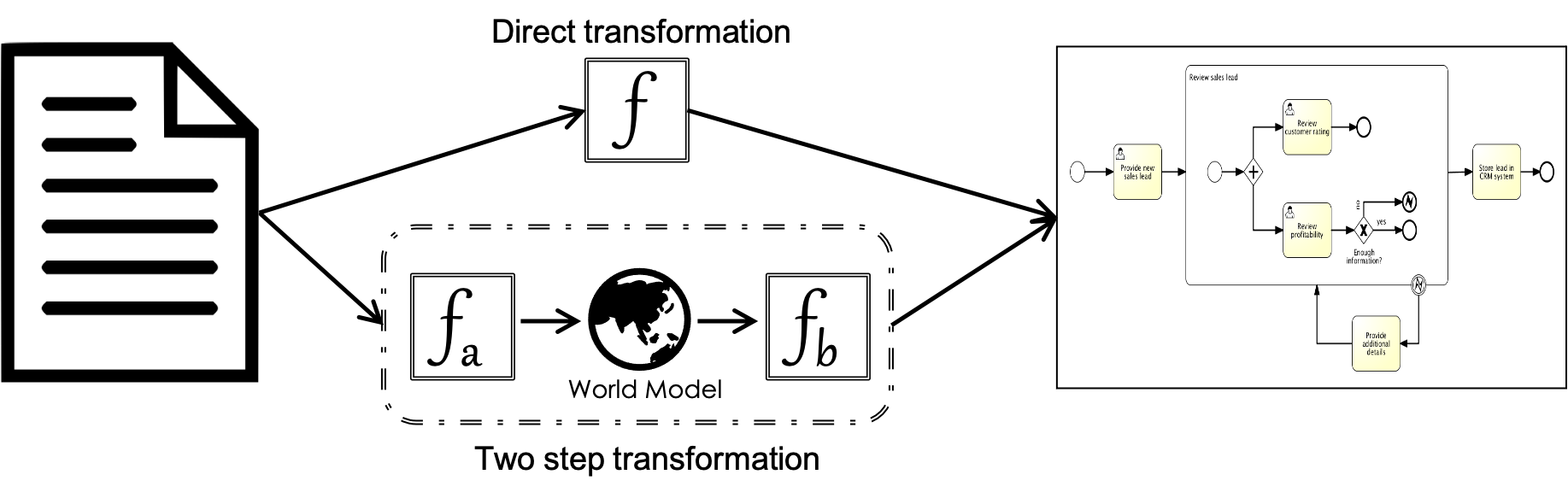}
    \caption{Two approaches to perform process extraction from text.}
    \label{img:texttomodelabstractview}
\end{figure}
If we look at the way this challenge has been tackled in the literature, we can roughly divide the approaches in two big categories. The first one aims to \textit{directly map} a process description into its process model representation via a single function $f$ as graphically depicted in the top part of Figure~\ref{img:texttomodelabstractview}\footnote{We have chosen BPMN as an illustrative example but the approach is clearly agnostic to the specific modeling language.}. In literature, function $f$ is typically implemented via a \textit{complex and ad-hoc tailored pipeline}. This approach has the advantage of defining a tailored transformation that can take into account all the available contextual information that can help solve the problem. Nonetheless, this advantage becomes a drawback when we need to devise general solutions or when the algorithmic function $f$ is applied into different contexts. A further approach towards the implementation of a direct mapping $f$ is the exploitation of Artificial Neural Networks. However, the huge quantity of data required to learn a model, and the small quantity of data available in this research domain makes this strategy rarely adopted.

The second approach found in the literature performs a \textit{two-steps transformation approach with intermediate representation} to extract and create a process model.
As illustrated in the bottom part of Figure~\ref{img:texttomodelabstractview} the algorithmic function $f$ is here considered a \textit{compound} function $\fa \circ \fb$: first, function $\fa$ extracts process elements from text and populates the intermediate representation (also called, world model), then function $\fb$ builds the process model diagram starting from the structured representation of the elements contained in the world model. Often $\fa$ and $\fb$ are further broken down into smaller tasks that allows to better handle the problem complexity. 

Similarly to other works (see Section~\ref{sec:related}), our investigation is performed building a computational function $f_a$, that aims at extracting key elements from text that can populate an intermediate structured representation, which could then be refined or used to build diagrammatic models.    
%
%
\subsection{In-context learning}
\label{ssec:prompting} 

Large language models (LLMs), such as \gpt (The Generative Pre-trained Transformer model)~\cite{Brown2020} or BERT~\cite{DBLP:conf/naacl/DevlinCLT19}, are pre-trained language models built by using an impressive amount of data and exploiting the advances of deep learning engineering and computational power~\cite{Brown2020,Raffel2020}. LLM are becoming a hot topic in Natural Language Processing (NLP) as they can be adopted, and fine-tuned, to solve complex tasks in different domains, such as open question answering in prototypical common-sense reasoning~\cite{DBLP:conf/emnlp/BoratkoLODLM20}.

While the fine tuning of LLMs for task-specific applications has become standard practice in NLP in the last few years, the advent of \gpt has pushed (or better, reduced) the amount of training to the extreme, suggesting a customisation by instructions and, if necessary, few shots (a.k.a. few examples) of the task to solve, without the need of updating the huge amount of parameters of the underlying model (that is, what is now called \textbf{in-context learning}). This approach has been shown to be extremely useful to address the training data challenge \cite{DBLP:conf/naacl/ScaoR21,DBLP:journals/corr/abs-2110-04374} and have been used to address topics ranging from medical dialogue summarization~\cite{pmlr-v149-chintagunta21a} to hate speech detection~\cite{DBLP:journals/corr/abs-2103-12407}.

In-context learning relying on the provision of instructions, some contextual knowledge\footnote{The terminology for this additional component varies from paper to paper.}, few examples, and the actual task to be solved that are coded into a single natural-language template called \textbf{prompt}. We illustrate the notion of prompt by showing an example of a one-shot fashion prompt template similar to the ones we used in the rest of the paper.

{\scriptsize
\begin{Verbatim}[breaklines=true,
				numbers=left, 
				xleftmargin=10mm, 
				frame=leftline, 
				numbersep=5mm]
  Considering the context of Business Process Management and process modelling and the following definitions: 
  Activity:
  An activity is a unit of work that can be performed by an individual or a group. It is a specific step in the process.
  Consider the following process:
  [ SAMPLE TEXT ]
  Q: lists the activities of the process
  A: [SAMPLE ANSWER]
  Consider the following process:
  [ ACTUAL TEXT ]
  Q: lists the activities of the process
  A: 
\end{Verbatim}
}

\noindent
Lines 1--3 provide the model with some contextual knowledge that is used to narrow the model's ``reasoning ability'' to the specific context at hand (BPM, in our case), and helps the model to disambiguate the possible different meaning of a word (e.g., activity). In our example they are composed by the identification of the domain (Business Process Management) and a definition.  
Lines 4--7 provide a one-shot example of the task at hand. In our case, a textual process description, the  \textit{task instructions} to be performed upon the text (that is, the specific query), and the correct answer.
Finally, lines 9 and 10 describe the actual problem to be solved, that is the text to be considered and the task to be performed upon the text. Line 11 ends the prompt by asking the model to produce an answer. 
At inference time, the prompt is inputed to the LLM in order to support the generation of the answer.

In this paper we investigate the adoption of in-context learning to overcome the lack of training data in process (information) extraction from text. In particular, we assess the usefulness of defining prompts that contain contextual knowledge in the form of conceptual definition of process model elements to be extracted, together with simple task instructions and two examples for each task.  
We also exploit the conversational nature of \gpt to investigate how to perform process extraction via incremental question answering. 

\subsection{The \PETD dataset}
\label{ssec:petd} 

\PETD \cite{DBLP:pet-dataset} is a novel dataset containing the human-annotated version of 45 textual descriptions corresponding to process models originally provided in~\cite{friedrich2010master}. Differently from the original work in \cite{friedrich2010master}, which provides pairs composed of a document and its corresponding (BPMN) model without any indication of which portion of text corresponds to which process element (part), \PETD provides specific annotations of the text with a number of process entities, and their relations. 
While the entire description of the dataset, the annotation guidelines, the annotation schema, and the annotation process are out of the scope of this paper\footnote{The intersted reader can refer to \cite{DBLP:pet-dataset} and to the entire dataset documentation that can be found at \url{https://w3id.org/pet}.} we report here the components of \PETD that are used in the paper.

\begin{wrapfigure}{l}{.45\textwidth}
  \centering
    \includegraphics[width=.45\textwidth]{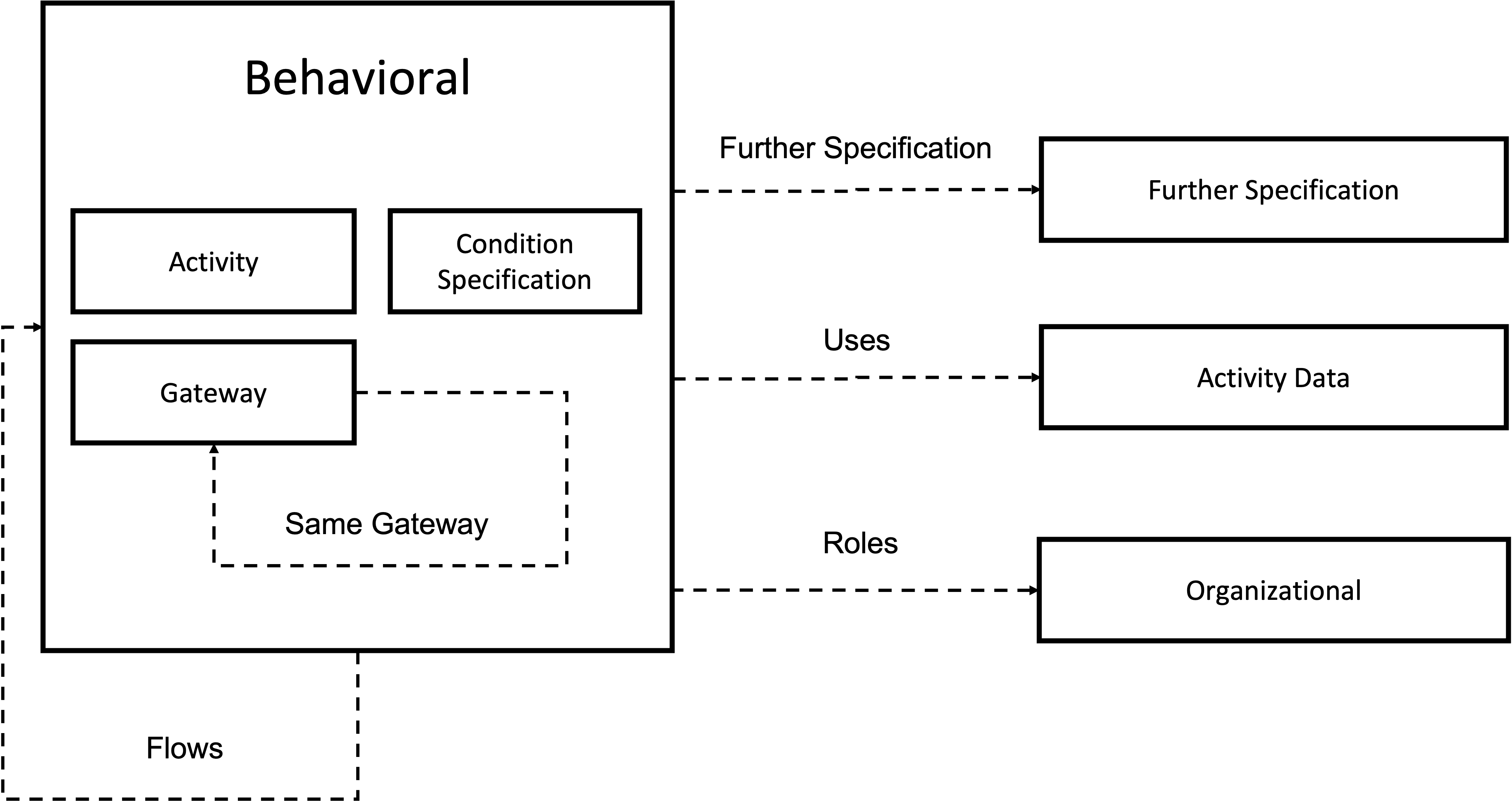}
  \caption{\PETD annotation schema.}
  \label{fig:figures_annotationschema}
\end{wrapfigure}
The articulated annotation schema of \PETD is reported in Figure \ref{fig:figures_annotationschema}. In this paper we make use of a concatenation of the ``activity''\footnote{The ``activity'' label is used in \PETD only to represent the verbal component of what is usually denoted as business process activity.} and ``activity data'' labels to identify activities in the text, of the organisational element connected with the ``performs'' role to an activity to identify performing actors and performs relations, and of the ``flow'' relation label to denote the (directly) follows relation. Since \PETD annotates within the  control flow also gateways and guards, we have considered here a simplified ``flow'' relation, implicitly subsumed by the annotations, that is obtained by directly connecting the appropriate activities and removing the extra control flow elements that we do not consider.  

\PETD also comes with extensive annotation guidelines that aim at levelling out different annotation styles of different people. These annotation guidelines contain definitions of different process elements and relationships. In our work we have exploited the definitions of \emph{flow} and \emph{sequence flow} to be used as contextual knowledge in prompts (See Section \ref{sec:approach-models}).

\section{Conversational process extraction from text via in-context learning}
\label{sec:approach}

We briefly describe here the approach we have followed to build different versions of conversational information seeking engines via in-context learning. 

Concerning the design of the conversational interaction, we have decided to focus on the extraction of \emph{activities}, actors performing an activity (hereafter \emph{participants}) (together with their \emph{performing} relation with activities), and \emph{directly follow} relations between activities (as in causal graphs), and \emph{performing} relation between actor and activity. We did focus on these four elements as they constitute somehow the basic building blocks on any business process model, and were therefore deemed an appropriate starting point for the empirical investigation of a new approach.    

Concerning the building of the conversational system, we have decided to start from \gpt (Generative Pre-trained Transformer 3) as pre-trained language model~\cite{Brown2020}  because it is the state-of-the-art standard in large  language models.
Since pre-trained language models are nor always suited for specific tasks, in their native form, we have decided to leverage in-context learning via prompting to provide different customizations of the native model.  

In the following we describe more in depth the strategies we have followed to build the series of questions, and the customized models that act as conversational systems. 

\begin{figure}[tb]  
	\centering
\scalebox{0.9}
{\begin{tcolorbox}
\begin{itemize}[parsep=3pt]
	\item[Q1:]\emph{Lists the activities of the process};
	\item[Q2:]\emph{Who is the participant performing activity X in the process model?} \\
	for each activity returned by Q1; 
	\item[Q3:]\emph{Considering the list of process activity described in the text, does activity X  immediately follow activity Y in the process model?}, \\
	for each pairs of different activities X and Y returned by Q1. 
\end{itemize}
    \end{tcolorbox}}\
    \caption{The questions and their ordering.}
    \label{img:questions}
\end{figure}

\subsection{Asking the questions}
\label{sec:approach-questions}
\begin{figure}[b]
  \centering	
    \includegraphics[width=.7\textwidth]{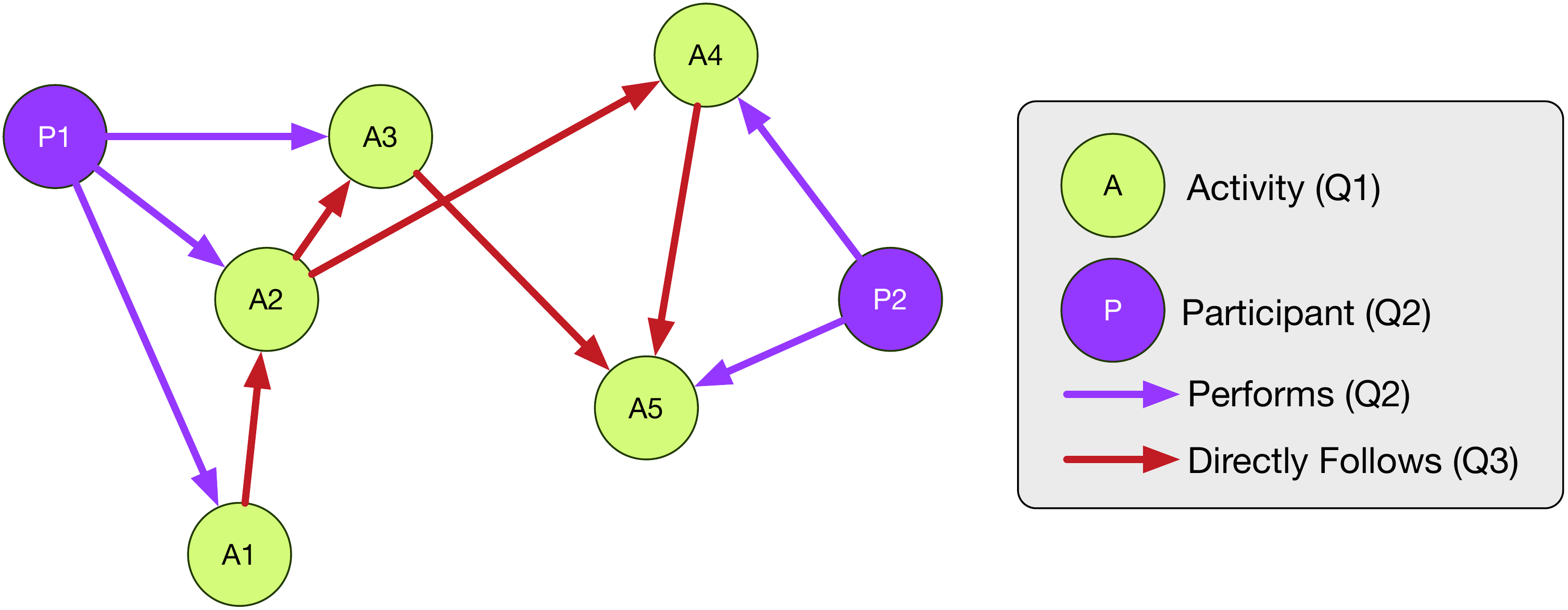}
  \caption{The elements and relations extracted.}
  \label{fig:figures_causalgraph}
\end{figure}

The questions used to extract information from text are reported in Figure~\ref{img:questions}. As it can be easily seen, these questions aim at building a structured representation such as the one represented in Figure~\ref{fig:figures_causalgraph}. The questions, and therefore construction, are posed in an incremental manner. In particular, first we ask questions about the process activities, then we enrich the activities with the actors performing them, and finally we ask about the precedence relation among activities.  

While the incremental order of the questions was somehow necessary, given the inability of the model to answer complex questions about the entire structure (e.g., ``list all the activities and their precedence relations'' - see Section \ref{sec:discussion}), this approach is also interesting because it mimics the way we manually build conceptual models (e.g., by interviewing domain experts with follow up questions) and because it enables the construction of flexible pipelines by combining different incremental questions.\footnote{Please note, that while asking all the questions manually can be cumbersome, incremental questions can be easily automatised to support specific pipelines.}.    
 
Another important aspect in setting up the questions is the specific wording to be used, that is, learning to ask the \emph{right} question. This first work didn't aim at investigating this aspect in depth. Nonetheless, Section~\ref{sec:discussion} reports some insights we gained in our work. 
 
\subsection{In-context learning customizations}
\label{sec:approach-models}
 
Our investigation relies on three different models. The first is the native pre-trained \gpt model. The further two are in-context learning customizations obtained by building two types of prompts that contain relevant contextual knowledge, and few shots (a.k.a. few examples) of the task to solve~\footnote{Towards reproducibility of the empirical assessment performed, the reader may find all prompts adopted at https://pdi.fbk.eu/cis/Prompts.zip}.

Contextual knowledge is provided by means of the identification of the specific domain (Business Process Management) and \emph{intensional definitions} of the process elements to be extracted. The advantage of relying on intensional definitions is indeed that of being compact and not having to rely on the provision of examples.  

\begin{figure}[tb]  
	\centering
\scalebox{0.8}
{\begin{tcolorbox}
Considering the context of Business Process Management and process modelling and the following definitions: 
    \begin{description}[parsep=3pt]
    	\item[Activity:] An activity is a unit of work that can be performed by an individual or a group. It is a specific step in the process.[\textbf{Q1},\textbf{Q2},\textbf{Q3}]
    	\item[Participant:] A participant is any individual or entity that participates in a business process. This could include individuals who initiate the process, those who respond to it, or those who are affected by it.[\textbf{Q2}]
	\item [Process Model:] A process model is a model of a process in terms of process activities and their sequence flow relations. [\textbf{Q3}]
    	\item[Flow:] A flow object captures the execution flow among the process activities. It is a directional connector between activities in a Process. It defines the activities’ execution order.[\textbf{Q3}]
     	\item[Sequence Flow:] A Sequence Flow object defines a fixed sequential relation between two activities. Each Flow has only one source and only one target. The direction of the flow (from source to target) determines the execution order between two Activities. A sequence relation is an ordered temporal relation between a source activity and the activity that immediately follow it in the process model.[\textbf{Q3}]
    \end{description}
    \end{tcolorbox}}\
    \caption{Instructions provided to the pre-trained model.}
    \label{img:definitions}
\end{figure}

The specific contextual knowledge contained in the prompts is depicted in Figure~\ref{img:questions}. They consist of a preamble identifying the BPM context, plus five definitions of Activity, Participant, Process Model, Flow and Sequence Flow. In the Figure each definition is labelled with the question it was used for. The definitions of Activity, Participant, and Process Model are slight rewordings of the answers obtained by asking the  question ``what is an activity (resp. participant, process model) in the context of Business Process Management?'' to \gpt itself. The definitions of Flow and Sequence Flow instead were extracted from the annotation instructions of the \PETD dataset that we use for the empirical assessment. These choices were made to minimise the external knowledge inserted in our assessment and - at the same time - provide a first empirical assessment of using intentional definitions in the customisation of pre-trained models.\footnote{Several definitions exist of many business process elements (see e.g., \url{https://www.businessprocessglossary.com}), but they often present different wordings and even conflicting characteristics~\cite{DBLP:conf/caise/AdamoFG20}. A thorough investigation of the impact of different definitions of business process elements in building conversational information seeking systems is out of the goal of this paper and is left for future works.}    

Few shot training was achieved by means of two samples consisting in two textual descriptions of processes together with the pairs of all questions and correct answers. The specific documents we did use as samples were documents \emph{2.2} 
 and \emph{10.9} 
 taken from the reference dataset provided by Friedrich in \cite{friedrich2010master}, while the answers were obtained from the \PETD annotated dataset \cite{DBLP:pet-dataset}.

\section{Empirical Assessment}
\label{sec:evaluation}

We would like to investigate the capability of pre-trained language models and in-context learning to extract different process elements from text. We hence aim at exploring the following research questions:

\begin{description}
\item[$\mathbf{RQ_1}$:] When a pre-trained language model is asked to extract specific information from a natural language process description, is it capable of answering appropriately?
\item[$\mathbf{RQ_2}$:] Are in-context learning approaches that rely on contextual BPM knowledge and few examples more effective in providing answers? Are there differences between providing contextual BPM knowledge and the examples? 
\end{description}

We provide below the process we performed to evaluate the proposed approach.
We start by introducing the data we used for preparing the evaluation, then we presents the adopted settings and the obtained results. Finally, we discuss what has been observed and the gathered insights.

\subsection{Datasets}
\label{ssec:datasets}

\begin{table}[t]
\centering
\renewcommand{\arraystretch}{1.2}
\scalebox{.8}{
\begin{tabular}{@{}l@{\quad}cccccc} 
\toprule
Text  & word\# & activity\# & participant\# & follow \# & perform\# \\
\midrule
T 1.2  & 100 & 10 & 2 & 10 & 10 \\
T 1.3 & 162 & 11 & 5 & 11 & 12 \\
T 3.3  & 71 & 7 & 2 & 6 & 4 \\
T 5.2  & 83 & 7 & 3 & 6 & 4 \\
T 10.1  & 29 & 4 & 2 & 4 & 4 \\
T 10.6  & 30 & 4 & 2 & 4 & 4 \\
T 10.13  & 39 & 3 & 2 & 2 & 3 \\
\bottomrule
\end{tabular}
}
\caption{\label{tab:dataset} The characteristics of the documents.}
\end{table}

The approach described in Section~\ref{sec:approach} has been empirically evaluated by performing a fine-grained analysis on \num documents extracted from the \PETD dataset illustrated in Section\ref{ssec:petd}. The number identifying the texts in the dataset and their characteristics in terms of number of words, and \PETD annotated elements activities, participants, performs and follows relations are reported in Table.


We are aware that the analysis of \num documents has limitations from a statistical significance point of view.
However, the rationale behind this empirical evaluation is two-fold.
First, since this is a first observational study of a promising groundbreaking strategy, we decided to select documents having specific characteristics in order to perform an ad-hoc analysis about how the pre-trained language model worked on them.
Second, the application of the proposed approach passed through several refinements round before to be tested since we had to understand how the pre-trained language model actually works.
Hence, to better understand the impact of the information provided by us to enrich the pre-trained language model, the most suitable way was to observe such behaviors on a small but characteristic subset of document.


In Section~\ref{sec:approach}, we mentioned the possibility of enhancing the pre-trained model by means of a few-shot strategy by starting from domain-specific documents.
In our experiments, we used the annotations provided by the documents $2.2$ and $10.9$ of the dataset.

\subsection{Experimental Setting}
\label{ssec:setting}
%
%
We evaluated the proposed prompting approach with four different settings:
\begin{description}
\item[\raw:] the GPT-3 model has been used as it is provided by the maintainers without any customization. This setting works as baseline to observe the capability of pre-trained language models of working within complex scenarios.
\item[\defs:] the GPT-3 raw model has been customized by providing the definitions shown in Figure~\ref{img:questions}. The aim was to inject domain-specific conceptual knowledge into the language model to observe the capability of the system to exploit the basic domain knowledge.
\item[\few:] the GPT-3 raw model has been enhanced by providing only few shots examples to trigger a quick train of the language model. In our case, such few shots were provided by using the gold annotations of the two documents mentioned in the previous subsection extracted from the \PETD dataset. Here, for the extraction of \textit{Activity} and \textit{Participant} only the annotations related to activities, activity data, and participants have been provided. While, for the extraction of \textit{Follows} and \textit{Performs} relationships, only the annotations related to the control flow and performing have been provided. This strategy has been adopted to avoid the injection of non-essential information that may cause noisy in the model.
\item[\both:] the GPT-3 raw model has been enhanced by using both strategies described above. 
\end{description}

Within all settings, the effectiveness of relation extraction has been observed by starting from the gold standard annotations and from the activities extracted from \textbf{Q1}.
This is an interesting comparison in order to observe if the usage of the extracted activities only has a detrimental effect on the extraction relationships or not.



\subsection{Results}
\label{ssec:results}

Table~\ref{tab:table-name2} provides the results of the empirical assessment performed on the portion of the \PETD dataset used for investigating the proposed approach.
The first column contains the document identifier.
The second column contains the elements or the relations for which the measures are provided (i.e., the extraction of the \textit{Activity}, \textit{Participant}, the \textit{Follow} relationship and the \textit{Performs} relationship)\footnote{ 
In few cases the model was able to provide semantically correct answers which did not match the exact \PETD labels. A paradigmatic case is the answer ``check and repair the computer'' as a single activity, instead of the two separate ones which are reported \PETD, as required by its specific annotation guidelines. We have carefully considered these few cases and decided to evaluate the semantically correct answers as correct answers.}.
Since the extraction of the relations \textit{Follow} and \textit{Performs} is influenced by the quality of the extraction of the activities, we did measure both the extraction of the relations for each activity returned as an answer to $\mathbf{Q_1}$, - suffix (ex) - and for each activity contained in the \PETD gold standard - suffix (gs). Thus, \emph{Follows (ex)} measures the ability of the system to extract the follows relation on the basis of the activities extracted by $\mathbf{Q_1}$, thus measuring the effective quality of the incremental Q/A interaction. Instead \emph{Follows (gs)} measures the ability of the system to extract the follows relation per-se on pairs of correctly identified activities, thus measuring the ability of the system on extracting the relation (similarly for \emph{performs}).      
Then, for each of the four settings described in subsection~\ref{ssec:setting}, we report the precision, recall and F1-score observed.


\begin{table}
\centering
\renewcommand{\arraystretch}{1.2}
\scalebox{.8}{
\begin{tabular}{@{}l@{\quad}lccc@{\qquad}ccc@{\qquad}ccc@{\qquad}ccc@{}} 
\toprule
& & \multicolumn{3}{c}{\raw} & \multicolumn{3}{c}{\defs} & 
\multicolumn{3}{c}{\few}  & \multicolumn{3}{c}{\both}  \\ 
\cmidrule(lr){3-5}\cmidrule(lr){6-8}\cmidrule(lr){9-11}\cmidrule(lr){12-14}
Text ID& Element & prec & rec & F1 & prec & rec & F1 & prec & rec & F1 & prec & rec & F1 \\
\midrule

\multirow{6}{*}{T 1.2} 
& Activity & \textbf{1.00} & 0.50 & 0.67 & 
			 \textbf{1.00} & 0.50  & 0.67 & 
			 0.75 & \textbf{0.90} & \textbf{0.82} &
			 0.75 & \textbf{0.90} & \textbf{0.82} \\
& Participant & \textbf{1.00} & \textbf{1.00} & \textbf{1.00} &
 			\textbf{1.00} & \textbf{1.00} & \textbf{1.00} &
 			\textbf{1.00} & \textbf{1.00} & \textbf{1.00} &
		 	\textbf{1.00} & \textbf{1.00}& \textbf{1.00} \\
		 	
& Follows (gs) &  0.00 & 0.00 & 0.00 & 
			0.33 & 0.10  & 0.15  & 
			\textbf{0.67} & \textbf{0.20} & \textbf{0.31} &
			0.40 & \textbf{0.20} & 0.27 \\
& Follows (ex) &  0.00 & 0.00 & 0.00 &
			 0.00 & 0.00 & 0.00  & 
			 \textbf{0.40} & 0.29 & 0.33 &
			 0.31 & \textbf{0.57} & \textbf{0.40} \\
& Performs (gs) &  0.20 & \textbf{1.00} & 0.33 &
			0.60 & \textbf{1.00}  & 0.75 & 
			\textbf{1.00} & \textbf{1.00} & \textbf{1.00} &
			\textbf{1.00} & \textbf{1.00} & \textbf{1.00} \\
& Performs (ex) &  0.33 & \textbf{1.00} & 0.50 & 
			\textbf{1.00} & \textbf{1.00}  & \textbf{1.00} &
			0.70 & \textbf{1.00} & 0.82 &
			0.70 & \textbf{1.00} & 0.82 \\

\midrule
\multirow{6}{*}{T 1.3} 
& Activity &  \textbf{1.00} & 0.69 & 0.82 & 
		\textbf{1.00} & 0.73  & 0.85 & 
		\textbf{1.00} & \textbf{0.87} & \textbf{0.93} &
		\textbf{1.00} & \textbf{0.87} & \textbf{0.93} \\	
& Participant & \textbf{1.00} & 0.60  & 0.75  &
		 \textbf{1.00} & 0.60  &  0.75 & 
		 \textbf{1.00} & \textbf{0.80} & \textbf{0.89} &
		\textbf{1.00} & \textbf{0.80}  &  \textbf{0.89} \\	
& Follows (gs) &  0.00 & 0.00 & 0.00 &
 		0.00 & 0.00  & 0.00  & 
 		 \textbf{0.18} & \textbf{0.73} & \textbf{0.29} &
	 	\textbf{0.18} & \textbf{0.73} & \textbf{0.29} \\
& Follows (ex) &  0.00 & 0.00 & 0.00 & 
 		0.13 & 0.07 & 0.09 & 
 		\textbf{0.22} & \textbf{0.75} & \textbf{0.34} &
		0.19 & 0.38 & 0.25 \\
	
& Performs (gs) &  \textbf{1.00} & \textbf{1.00} & \textbf{1.00} & 
		0.91 & 0.91 & 0.91 & 
		0.97 & \textbf{1.00} & 0.98 &
		0.97 & \textbf{1.00} & 0.98 \\
		
& Performs (ex) &  0.78 & \textbf{1.00} & 0.88 &
		0.82 & \textbf{1.00} & 0.90 & 
		\textbf{0.92} & \textbf{1.00} & \textbf{0.96} &
 		\textbf{0.92} & \textbf{1.00} & \textbf{0.96} \\

\midrule
\multirow{6}{*}{T 3.3} 
& Activity &  \textbf{1.00} & 0.57 & 0.73 & 
		0.86 & \textbf{1.00}  & 0.92 & 
		\textbf{1.00} & 0.86 & 0.92 &
		0.88 & \textbf{1.00} & \textbf{0.93}    \\
& Participant & \textbf{0.67} & \textbf{1.00} & \textbf{0.80} &
		0.50 & \textbf{1.00} & 0.67 &
		\textbf{0.67} & \textbf{1.00} & \textbf{0.80} &
		\textbf{0.67} & \textbf{1.00} & \textbf{0.80} \\
		
& Follows (gs) &  0.00 & 0.00 & 0.00 &
		 0.00 & 0.00 & 0.00 &
		\textbf{0.16} & \textbf{0.50} & \textbf{0.24} &
		\textbf{0.16} & \textbf{0.50} & \textbf{0.24} \\
& Follows (ex) &  0.00 & 0.00 & 0.00 &
		0.00 & 0.00 & 0.00 &
		\textbf{0.24} & \textbf{0.80} & \textbf{0.36} &
		0.15 & 0.67 & 0.25  \\
& Performs (gs) &  \textbf{0.57} & \textbf{1.00} & \textbf{0.73} &
		\textbf{0.57} & \textbf{1.00} & \textbf{0.73} &
		\textbf{0.57} & \textbf{1.00} & \textbf{0.73} &
		\textbf{0.57} & \textbf{1.00} & \textbf{0.73}  \\
& Performs (ex) & \textbf{0.75} & \textbf{1.00} & \textbf{0.86} &
		0.43 & \textbf{1.00} & 0.60 &
		0.67 & \textbf{1.00} & 0.80 &
		0.50 & \textbf{1.00} & 0.57  \\
		
\midrule
\multirow{6}{*}{T 5.2} 
& Activity &  0.00 & 0.00 & 0.00 & 
		\textbf{1.00} & 0.57  & 0.73 & 
		\textbf{1.00} & \textbf{0.86} & \textbf{0.92} &
		\textbf{1.00} & \textbf{0.86} & \textbf{0.92} \\
& Participant & 0.00 & 0.00 & 0.00 &
		\textbf{1.00} & \textbf{1.00} & \textbf{1.00} &
		\textbf{1.00} & \textbf{1.00} & \textbf{1.00} &
		\textbf{1.00} & \textbf{1.00} & \textbf{1.00}  \\
		
& Follows (gs) &  0.00 & 0.00 & 0.00 & 
		0.00 & 0.00  & 0.00  &
		0.23 & \textbf{0.83} & 0.36 &
		\textbf{0.24} & \textbf{0.83} & \textbf{0.37} \\
& Follows (ex) &  0.00 & 0.00 & 0.00 &
		 0.00 & 0.00 & 0.00  & 
		 0.24 & \textbf{0.80} & 0.36 &
		 \textbf{0.25} & \textbf{0.80} & \textbf{0.38} \\
		 
& Performs (gs) &  \textbf{0.57} & \textbf{1.00} & \textbf{0.73} &
 		\textbf{0.57} & \textbf{1.00} & \textbf{0.73} &
 		0.48 & \textbf{1.00} & 0.64 &
 		0.43 & \textbf{1.00} & 0.60 \\
& Performs (ex) &  0.00 & 0.00 & 0.00 &
		\textbf{0.75} & \textbf{1.00}  & \textbf{0.86} &
		0.39 & \textbf{1.00} & 0.56 &
		 0.33 & \textbf{1.00} & 0.50 \\

\midrule
\multirow{6}{*}{T 10.1} 
& Activity &  0.00 & 0.00 & 0.00 &
		0.00 & 0.00 & 0.00 &
		\textbf{1.00} & \textbf{1.00} & \textbf{1.00} &
		\textbf{1.00} & \textbf{1.00} & \textbf{1.00} \\
& Participant & 0.00 & 0.00 & 0.00 &
 		0.00 & 0.00 & 0.00 &
 		\textbf{1.00} & \textbf{1.00} & \textbf{1.00} &
 		\textbf{1.00} & \textbf{1.00} & \textbf{1.00}  \\
& Follows (gs) &  0.00 & 0.00 & 0.00 &
		0.00 & 0.00  & 0.00  &
		\textbf{0.50} & \textbf{1.00} & \textbf{0.67} &
		0.38 & \textbf{1.00} & 0.55 \\
& Follows (ex) &  0.00 & 0.00 & 0.00 & 
 		0.00 & 0.00 & 0.00  & 
 		0.33 & \textbf{1.00} & 0.50 &
 		\textbf{0.43} & \textbf{1.00} & \textbf{0.60} \\
& Performs (gs) &  \textbf{1.00} & \textbf{1.00} & \textbf{1.00} &
		\textbf{1.00} & \textbf{1.00} & \textbf{1.00} &
		\textbf{1.00} & \textbf{1.00} & \textbf{1.00} &
		\textbf{1.00} & \textbf{1.00} & \textbf{1.00} \\
& Performs (ex) &  0.00 & 0.00 & 0.00 &
		0.00 & 0.00 & 0.00 &
		\textbf{1.00} & \textbf{1.00} & \textbf{1.00} &
		\textbf{1.00} & \textbf{1.00} & \textbf{1.00} \\
		
\midrule
\multirow{6}{*}{T 10.6} 
& Activity &  0.00 & 0.00 & 0.00 &
		 0.00 & 0.00 & 0.00 &
		 \textbf{1.00} & \textbf{1.00} & \textbf{1.00}  &
		 \textbf{1.00} & \textbf{1.00} & \textbf{1.00} \\
& Participant & 0.00 & 0.00 & 0.00 &
		 0.00 & 0.00 & 0.00 &  
		\textbf{1.00} & \textbf{1.00} & \textbf{1.00} &
		\textbf{1.00} & \textbf{1.00} & \textbf{1.00}  \\
& Follows (gs) & 0.00  & 0.00 & 0.00 &
		 0.00  & 0.00 & 0.00  &
		 \textbf{0.60} & \textbf{1.00} & \textbf{0.75} &
		 0.43 & \textbf{1.00} & 0.60  \\
& Follows (ex) &  0.00 & 0.00 & 0.00 &
		0.00 & 0.00 & 0.00  &
		\textbf{0.60} & \textbf{1.00} & \textbf{0.75} &
		0.43 & \textbf{1.00} & 0.60 \\
& Performs (gs) &  \textbf{1.00} & \textbf{1.00} & \textbf{1.00}  &
		\textbf{1.00} & \textbf{1.00} & \textbf{1.00}  &
		\textbf{1.00} & \textbf{1.00} & \textbf{1.00} &
		\textbf{1.00} & \textbf{1.00} & \textbf{1.00}  \\
& Performs (ex) &  0.00 & 0.00 & 0.00 &
		0.00 & 0.00 & 0.00  &
		\textbf{1.00} & \textbf{1.00} & \textbf{1.00} &
		\textbf{1.00} & \textbf{1.00} & \textbf{1.00} \\

\midrule
\multirow{6}{*}{T 10.13} 
& Activity &  0.00 & 0.00 & 0.00 & 
		0.00 & 0.00 & 0.00 &
		\textbf{1.00} & \textbf{1.00} & \textbf{1.00} &
		0.60 & \textbf{1.00} & 0.75 \\
& Participant & 0.00 & 0.00 & 0.00 &
		0.00 & 0.00 & 0.00 & 
		\textbf{1.00} & \textbf{1.00} & \textbf{1.00} &
		\textbf{1.00} & 0.5 & 0.67  \\
& Follows (gs) & \textbf{0.67}  & \textbf{1.00} & \textbf{0.80} &
		0.00 & 0.00  & 0.00 &
		0.29 & \textbf{1.00} & 0.44 &
		0.25 & \textbf{1.00} & 0.40 \\
& Follows (ex) &  0.00 & 0.00 & 0.00 &
		0.00 & 0.00 & 0.00  &
		\textbf{0.29} & \textbf{1.00} & \textbf{0.44}  &
		0.22 & \textbf{1.00} & 0.36 \\
& Performs (gs) &  \textbf{1.00} & \textbf{1.00} & \textbf{1.00} &
		\textbf{1.00} & \textbf{1.00} & \textbf{1.00} &
		0.67 & \textbf{1.00} & 0.80 &
		0.67 & \textbf{1.00} & 0.80 \\
& Performs (ex) &  0.00 & 0.00 & 0.00 &
		0.00 & 0.00 & 0.00  &
		\textbf{1.00} & \textbf{1.00} & \textbf{1.00} &
		0.40 & \textbf{1.00} & 0.57 \\
\bottomrule
\multirow{6}{*}{\textbf{Average}} 
& Activity & 0.43 & 0.25 & 0.32 & 
		0.55 & 0.40 & 0.45 &
		\textbf{0.96} & 0.93 & \textbf{0.94} &
		0.89 & \textbf{0.95} & 0.91  \\
		
& Participant & 0.38 & 0.37 & 0.36 &
		0.50 & 0.51 & 0.49 &
		\textbf{0.95} & \textbf{0.97} & \textbf{0.96} &
		\textbf{0.95} & 0.90 & 0.91\\
		
& Follows (gs) &  0.10 & 0.14 & 0.11&
		0.05 & 0.01 & 0.02 &
		\textbf{0.38} & \textbf{0.75} & \textbf{0.44} &
		0.29 & \textbf{0.75} & 0.39 \\
		
& Follows (ex) & 0.00 & 0.00 & 0.00 &
		0.02 & 0.01 & 0.01 &
		\textbf{0.33} & \textbf{0.81} & \textbf{0.44} &
		0.28 & 0.77 & 0.41 \\
		
& Performs (gs) & 0.76 & \textbf{1.00} & 0.83 &
		\textbf{0.81} & 0.99 & 0.87 &
		\textbf{0.81} & \textbf{1.00} & \textbf{0.88} &
		\textbf{0.81} & \textbf{1.00} & 0.87\\
& Performs (ex) & 0.27 & 0.43 & 0.32 &
		0.43 & 0.57 & 0.48 &
		\textbf{0.81} & \textbf{1.00} & \textbf{0.88} &
		0.69 & \textbf{1.00} & 0.77\\
\bottomrule
\end{tabular}
}
\caption{\label{tab:table-name2} The results for the four settings.}
\end{table}

The results obtained highlight few interesting patterns.
In general, we can see that strategy \raw provides unsatisfactory results, thus highlighting that the GPT-3 raw model is not able to address the task of extracting different process elements from text n a satisfactory manner, and answering ``NO'' to $\mathbf{RQ_1}$. Few observations can nonetheless be made. Especially in texts 1.2, 1.3, and 3.3
\raw provides good results for the extraction of participants and the performs (gs) relationship. 
This may show that information about actors performing something is somehow part of the general knowledge that a pre-trained linguistic model has.   
Texts 1.2, 1.3 and 3.3, \raw provides also acceptable results for activity, thus highlighting that the \raw model works well with complex texts.
This may be a consequence of the fact that longer texts provide a more complete context of the process description.
Hence, such a context better helps the \raw model in understanding which are the most relevant elements to detect.

Concerning $\mathbf{RQ_2}$ we can see that in-context learning approaches that rely on contextual BPM knowledge and few examples more effective in providing answers than the native \gpt model, and can lead to good results. In particular \few appears to be the best overall strategy. Adding contextual BPM knowledge, and in particular definitions, is useful in specific cases - thus hinting to a possible positive effect - but does not appear to be a valid general strategy, neither when it is provided alone (as in \defs) not when it complements the examples (as in \both). Whether this is the effect of the definitions we tested or definitions in general is left for future work to assess. The result we can report here is that adding the \emph{right} contextual BPM knowledge may be a non trivial problem that needs to be carefully investigated. 
A exception to the overall satisfactory performance of \few is given by the \textit{Follow} relationship both in its (ex) and (gs) settings. In fact, while providing the two examples is useful to increase the performances w.r.t. the \raw baseline, the gain is often limited. This result may indicate that the knowledge of temporal relationships of \gpt is insufficient for the BPM domain and better ad-hoc training in needed. To sup up, we can positively answer $\mathbf{RQ_2}$ saying that in-context learning in a few-show fashion can prove beneficial to extract process information from text with the need to better investigate the follows relationship in the future.   

Finally, we observed how the performance are not related to the length of the documents.
Indeed, from Table~\ref{tab:dataset}, we may observe that we worked with long documents (i.e., Texts 1.2, and 1.3), medium-size documents (i.e., Texts 3.3 and 5.2), and short documents (i.e., Texts 10.1, 10.6 and 10.13).
By observing the behavior of each setting reported in Table~\ref{tab:table-name2}, no relevant differences may be observed from the metrics.
This aspect is particularly interested since it means that the proposed strategy may be applied in different scenarios without considering the documents length as a criticality to address.
\section{Discussion}
\label{sec:discussion}

This work represents a first attempt towards the use of in-context learning techniques to enable the usage of conversational information seeking strategies for building business process models in an incremental way.
From this experience we gathered several insights, that encompass the results reported in Table~\ref{tab:table-name2}. We report them below as they will trigger future investigation.
We may group such insights within three main categories: (i) the type of interaction between the user and the model; (ii) the parameters to adopt for understanding the behavior of the model; and, (iii) the understanding of which information better benefits the effectiveness of the in-context learning.

\textit{Interaction between the user and the model.}
The first important aspect to consider is how the interaction between the user and the machinery occurs, i.e., how the content of conversations is structured.
It has to be as much simple, and complete, as possible.
Indeed, the model (independently by the adopted setting described in subsection~\ref{ssec:setting}) is not able to provide proper answers to generic and complex questions.
An example is the following: \textit{List all the activities and their precedence relations.}
Here, two information are requested at the same time and having the second one that is semantically dependent by the first one.
We observed that this type of requests led to an empty or, in general, not significant outcome.
The same issue occurs when questions are not complete.
For example, if we omit the word \textit{performer} when we ask for who is the actor performing a specific activity, the model tends to generate a wrong answer.
This is an important aspect to take into account since the model has a limited inference capability with respect to the human-like cognitive process.

\textit{Parameters to understand model behavior.}
During preliminary tests, i.e., before to run the empirical assessment, we tested different prompt templates.
This type of test was necessary to observe if the format of the template may affect the quality and variety of the results.
This preliminary investigation confirmed this hypothesis since, depending on the prompt format, results are different.
Similarly, we investigated the importance of model parameters. 
By changing the sampling strategy, the model effectiveness changed as well. 
In particular, the augmentation of model \textit{creativity}, i.e., the way with which the textual output is generated, may have relevant detrimental effects. 
A creative setting is useful in preliminary investigations, especially when we explored which knowledge was already available within the model about a particular topic.
The model \textit{creativity}~\footnote{The model creativity is set by two parameters that set the sampling strategy: \textit{temperature} and \textit{nucleus}.}
 is encoded through a numerical value ranging from $0.0$ to $2.0$.
\textit{Creativity} could help in solving complex reasoning tasks.
But, the more the \textit{creativity} value is increases, the lower could 
 be the similarity between generated textual outputs by starting from the same question but requested at different timestamps.
For this reason, in order to make experiments reproducible, the \textit{creativity} value has to be set to $0.0$.
%

\textit{Understanding of which information better benefits the effectiveness of the in-context learning.}
The third aspect relates to the selection of information to provide for the in-context learning phase in order to improve the model capability in answering correctly.
A straightforward aspect we validated is that the choice of examples used for the few-shot customization has a relevant impact on the results. 
During the preliminary tests mentioned in the previous paragraph, we chosen different textual descriptions to observe the inference capability of the system in order to tune the queries to perform during our empirical assessment.
As example, the use of the term \textit{Participant} instead of \textit{Actor} led to better results.
The hypothesis is that the term \textit{Actor} may be considered a more general term used in several contexts.
While, the word \textit{Participant} brings the semantic meaning of someone having an active role in a task.
This aspect will be further investigated in the future.
Another case was instead related to ambiguous knowledge that may be previously loaded into the model.
Indeed, we observed that, concerning the concept of \textit{Activity} different definitions of such a concept were already included in the model.
Hence, in order to preserve the capability of acting in an effective way within the business process domain, we adopted the prefix text \textit{Consider the context of Business Process Management and Process modeling} before queries related to the extraction of activities from the text.
This trigger a higher effectiveness of the model.

\textit{Potential impact on annotation campaigns.}
The performance obtained by the proposed approach highlight how in-context learning techniques may be used side-by-side with human experts within annotation campaigns~\cite{DBLP:conf/emnlp/WangLXZZ21}.In general, the annotation of natural language texts describing business process models is a complicated task to be performed from scratch.
A potential impact of approaches like the ones discussed in this paper is to support the annotation task by providing candidate annotations related to the process elements for which the performance are acceptable, e.g. \textit{Participants} and \textit{Performs} relationships.
Then, it would be possible to demand the human experts the detection of more complex process elements, e.g. \textit{Control flows}.
This is a preliminary analysis of this aspect: future investigations will consider how to integrate in the most effective way these two annotation ways.

The analysis we performed on the behavior of the pre-trained model and on the in-context learning customization paved the way to further explore the approached proposed in this paper.
The outcomes of the performed empirical assessment demonstrated the suitability of the proposed approach that has the potential of becoming a groundbreaking strategies for the expert-supervised incremental building of business process models from texts by exploiting the conversational information seeking paradigm.

\section{Related Work}
\label{sec:related}

We can group existing work on extracting process elements or an entire process model from text in three different streams.

In the first stream, process extraction from text is addresses trough a direct mapping function. \cite{AckermannV13} proposes to learns how to build up a process model diagram through user'
s feedbacks. Whenever a sentence is provided in input to the tool, the engine checks if a corresponding mapping rule exist. If there are no rules to describe a process element, the system exploits the user'
s feed-backs to create one. So, the the user has to manually map a text fragment into its process model element. Then, the system creates the corresponding mapping rule. The works of \cite{derAaDiCiccio19} and \cite{LopezDHM18} target the extraction of declarative constraints expressed in DECLARE and DCR graphs respectively in a rule based fashion. The extraction is performed from single sentences and focuses on the identification of roles, activities, and a specific set of DECLARE (resp. DCR graph) constraints. Finally, the recent work of \cite{Han2020} proposes a deep learning solution to the problem of automatically
discovering a business process and its corresponding process model diagram (represented in BPMN) without extra human labeled knowledge. The generality of this approach is hampered by the restriction of the input to heavily structured Process Definition Documents.   	  

The second stream contains works that address process extraction from text as a two-steps transformation approach with intermediate representation
\cite{Friedrich11,GoncalvesSB11,SawantRSPS14,EpureMHDS15,HonkiszK018}. All these works produce BPMN process model diagrams by exploiting rules or templates. The seminal work of \cite{Friedrich11} exploits templates and a slight modification of a CREWS intermediate World Model to extract BPMN diagrams from text without making assumptions regarding the input text. It also introduces a publicly available dataset of 47 pairs composed of a process model description, and a process model diagram for performing the evaluation, which is still considered as one of the reference datasets up-to-date. Despite still being one of the reference works on this topic, the work in \cite{Friedrich11} failed to address the variability of real world texts and was not developed further. \cite{GoncalvesSB11} tackles the challenge of process extraction from text  by proposing a \emph{story telling mining} method to ease the work of analysts by automatically eliciting models from stories. \cite{SawantRSPS14} presents a method to create a process model diagram and its associated ontology from semi-structured use case descriptions (descriptions on how an entity should cooperate with other entities), while \cite{EpureMHDS15} focuses on semi-structured textual descriptions of process executions in the archaeological domain. Finally, \cite{HonkiszK018} exploits a syntactic analysis of the input text to extract Subject-Verb-Object constructs and keywords-based extraction for the identification of Gateways. 

The third stream of work instead tackles only the $f_a$ function of the two-steps transformation approach, thus extracting information as lists of tags from a text. In particular, the works in \cite{Ferreira17} and \cite{DBLP:conf/bpm/QuishpiCP20} follow a rule based approach to extract mainly control and data flows in a procedural and declarative fashion, respectively. \cite{Leopold18} exploits a support vector machine (SVM) approach in order to classify manual tasks, user tasks, and automated tasks in the context of Robotic Process Automation. Finally, the recent work of  \cite{DBLP:conf/caise/0003W0LLZLW20} proposes a hierarchical neural network approach, called Multi-Grained Text Classifier (MGTC), to tackle the problem of classifying sentences in procedural texts without engineering any features. Unfortunately, the approach is only applied to sequential lists of tasks such as recipes or assembly instructions. 

Further works, such as the ones in \cite{DBLP:journals/is/AaLR17} and \cite{DBLP:journals/sosym/Sanchez-Ferreres21}, aim at supporting different forms of alignments between textual descriptions and business process models, and are therefore not considered here.     

In the context of process extraction from text, our works belongs to the third stream above. Differently from state of the art research, and to the best of our knowledge, this is the first research endeavour aimed at extracting process information from text using a deep learning method based on pre-trained language representation models which aims at dealing with an entire textual description without making assumptions regarding the input text. It also does it in an incremental and flexible conversational fashion so as to extract the required information via question and answering dialogues. 
\section{Conclusion}
\label{sec:conclusion}

In this paper we explored the feasibility of using in-context learning over the pre-trained language models to perform process extraction from textual documents in an incremental question and answering manner. In particular we explored the feasibility of using the native \gpt model, and two customizations built by providing conceptual definitions of business process elements and a limited number of examples in a few shot learning fashion. 
The results highlighted the potential of the in-context learning approach which can substantially address the ``training data challenge'' of deep learning based NLP techniques the BPM field.
The results obtained by the in-context learning strategy opened the possibility to use this technique to address the construction of business process model by starting from natural language text in scenarios where it is necessary to manage the low-resources issues and by exploiting the human-in-the-loop paradigm given the role of the domain expert in processing the information provided by the model.
We reported a suite of lessons learned from this experience that will drive the development of further research questions.



%
%
%
\bibliographystyle{splncs04}
 \bibliography{bpm2022GPT3}
\end{document}